\documentclass[letterpaper, 10 pt, conference]{ieeeconf}  
\usepackage{graphicx}
\usepackage[linesnumbered,ruled,vlined]{algorithm2e}
\usepackage{algorithmicx}
\usepackage{amsmath}
\usepackage{mathtools}
\usepackage{booktabs}
\usepackage{array}
\usepackage{subcaption}
\usepackage{booktabs}
\usepackage{tabularx}
\usepackage{multirow}
\usepackage{threeparttable}
\usepackage{float}
\usepackage{svg}
\usepackage{amssymb}
\usepackage{amsfonts}
\usepackage{float}
\usepackage{comment}
\usepackage{url} 
\usepackage{balance}
\usepackage{cite}
\usepackage{multicol}
\usepackage[noend]{algpseudocode}
\usepackage[colorlinks=true, linkcolor=red, citecolor=red, urlcolor=red]{hyperref}

\IEEEoverridecommandlockouts                              

\title{\LARGE \bf
SeGMan: Sequential and Guided Manipulation Planner for Robust Planning in 2D Constrained Environments
}

\author{Cankut Bora Tuncer$^{1,\dagger}$, Dilruba Sultan Haliloglu$^{1,\dagger}$, and Ozgur S. Oguz$^{1}$
\thanks{$^{\dagger}$These authors contributed equally to this work.}
\thanks{$^{1}$Dept. of Computer Engineering, Bilkent University.}
\thanks{*This work was supported by TUBITAK under 2232 program with project number 121C148 (``LiRA").}
\thanks{\textbf{Corresponding author:} Cankut Bora Tuncer, Department of Computer Engineering, Bilkent University, 06800 Bilkent, Ankara, Turkey. Email: {\tt\small bora.tuncer@bilkent.edu.tr}}%
}

\begin{document}

\maketitle
\thispagestyle{empty}
\pagestyle{empty}

\begin{abstract}
    In this paper, we present SeGMan, a hybrid motion planning framework that integrates sampling-based and optimization-based techniques with a guided forward search to address complex, constrained sequential manipulation challenges, such as pick-and-place puzzles. SeGMan incorporates an adaptive subgoal selection method that adjusts the granularity of subgoals, enhancing overall efficiency. Furthermore, proposed generalizable heuristics guide the forward search in a more targeted manner. Extensive evaluations in maze-like tasks populated with numerous objects and obstacles demonstrate that SeGMan is capable of generating not only consistent and computationally efficient manipulation plans but also outperform state-of-the-art approaches. \url{https://sites.google.com/view/segman-lira/}
\end{abstract}

\section{Introduction}

Sequential manipulation is used in complex and constrained environments where tasks require multiple steps. It involves generating a feasible motion plan by decomposing a primary goal into manageable subgoals that can be addressed individually. However, in constrained environments, finding suitable subgoals is a challenging process that demands the identification of critical objects and placement locations. Due to the high dimensionality of the environments, the subgoal generation process requires coordinated reasoning through multiple stages of interaction, such as grasping, reorienting, pushing, and removing objects, in a structured manner. However, even minor errors, such as a misaligned object, can lead to failures in subsequent steps, necessitating re-planning and extending plan generation time. Thus, the quality of the generated subgoals not only directs the motion planning process but also impacts the robustness and efficiency of the final solution.



Several studies have investigated methods for identifying and addressing critical subgoals in sequential motion planning. 
\cite{levit2024solvingsequentialmanipulationpuzzles} proposes a heuristic-driven search to identify intermediate subgoals, which are then solved using Task-and-Motion Planning (TAMP). However, the heuristics lack generalization, often misguiding the search process and increasing computation time.
A different approach is taken in~\cite{cicek2024hmapiterativehybridsequential}, where the subgoals are generated with a sampling-based method and utilized in motion plan optimization. While effective, this method struggles to propose feasible obstacle relocation points and has limited adaptability to varying environmental factors.
In~\cite{rearrangement_puzzle, zhang2024learn2decomposelearningproblemdecomposition}, novel subgoal generation algorithms are proposed to decompose complex problems into simpler, more manageable subgoals.
However, a common limitation among these approaches is that their proposed planners lack the capability for fine manipulation of objects in constrained environments. 
Thus, current methods do not offer a framework capable of identifying critical subgoals for constrained sequential manipulation while offering robustness and computational efficiency.

\begin{figure}[!t]
    \centering
    \resizebox{0.45\textwidth}{!}{\includegraphics{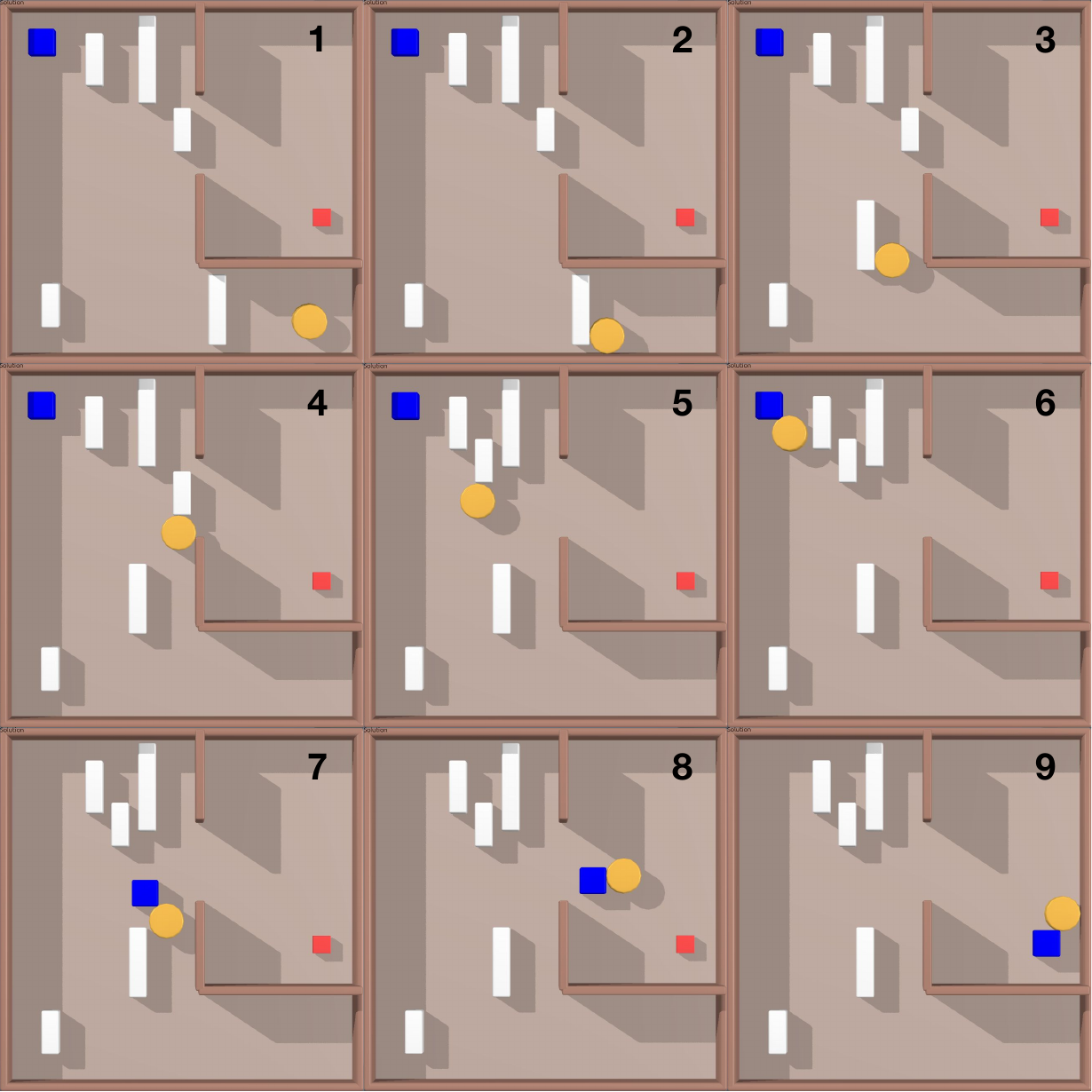}}
    \caption{The example illustrates a pick-and-place task where the agent (yellow) must move the goal object (blue) to the goal (red). Before doing so, it selectively removes the obstacles (white) to ensure it can access the goal object.}
    \label{fig:intro}
\end{figure}

Fig.~\ref{fig:intro} illustrates a simplified task where a 2DOF agent (yellow cylinder) must execute sequential pick-and-place actions to move the goal object (blue square) to its goal (red square) by interacting with movable obstacles (white squares)~\cite{levit2024solvingsequentialmanipulationpuzzles}. 
The solution of the problem requires relocation of two critical movable obstacles (Fig.~\ref{fig:intro}, 1-5) so that a feasible motion plan can be realized from agent to goal object (Fig.~\ref{fig:intro}, 6) and goal object to the goal (Fig.~\ref{fig:intro}, 7-9). 
This problem highlights the key challenges in sequential manipulation, including effective subgoal selection and precise object manipulation to iteratively solve each subgoal.

\setlength{\belowcaptionskip}{-10pt}


In this paper, we introduce SeGMan — Sequential and Guided Manipulation Planner — a hybrid motion planner that integrates sampling and optimization-based techniques with a heuristic-driven forward search. By using waypoints from a sampling-based planner as subgoals~\cite{cicek2024hmapiterativehybridsequential}, SeGMan adaptively selects the subgoals to optimize object manipulation, which enhances efficiency across various constrained environments. Moreover, the proposed generalizable forward search heuristics guide the search on critical obstacles and relocation positions, improving robustness.

Our main contributions are as follows:
\begin{itemize}
    \item A generalizable hybrid motion planning framework by introducing a guided search for obstacle manipulation in constrained environments.
    \item A novel algorithm that identifies critical obstacles and leverages generalizable heuristics to propose and evaluate obstacle relocation subgoals for guided forward search in scenarios where unguided search would be impractical.
    \item A method that adaptively selects subgoals and iteratively adjusts granularity, improving the robustness and efficiency of the framework.
\end{itemize}

The framework was evaluated across 15 distinct problems, similar to the task in Fig.~\ref{fig:intro}, each presenting unique challenges with varying levels of complexity.
Overall, SeGMan outperformed baseline methods in robustness and demonstrated a balance between computation time and solution quality.

\section{Related Work}
Motion planning for object manipulation involves determining a sequence of collision-free and feasible configurations for both the object and the agent, from the initial state to the goal state. This is typically achieved using optimization-based, sampling-based, or hybrid approaches.

\subsection{Optimization-Based Methods}
The traditional optimization-based methods optimize trajectories for smoothness and efficiency. While CHOMP ~\cite{zucker2013chomp,dragan2011manipulation,dragan2017learning} and variants rely on gradient-based optimization, STOMP~\cite{kalakrishnan2011stomp} uses stochastic sampling to explore multiple trajectories. TrajOpt~\cite{schulman2014motion} introduces convex collision avoidance constraints, and KOMO~\cite{14-toussaint-KOMO} extends trajectory optimization by incorporating higher-order kinematic constraints and leveraging second-order Newton methods. Although optimization-based methods offer computational efficiency, they are limited to finding locally optimal solutions, reducing their effectiveness in high-dimensional, long-horizon tasks. 

\subsection{Sampling-Based Methods}
In sampling-based manipulation planning, the typical approach involves the exploration of high-dimensional spaces by randomly sampling predefined primitives.
CBiRRT~\cite{berenson2009manipulation} employs grasping primitives, RMR*~\cite{schmitt2017optimal} relies on constraint manifold-based motion primitives, and IMACS~\cite{kingston2019exploring} utilizes implicit constraint-compliant primitives.
While these primitives help guide the sampling process, their predefined nature limits generalizability.
To address this limitation, CMGMP~\cite{cheng2022contact} introduces an approach that automatically generates motion primitives to enhance adaptability.
However, despite these advancements, sampling-based methods generally require further trajectory optimization and remain computationally expensive in complex environments.

\subsection{Hybrid Methods}
Hybrid methods utilize both sampling and optimization-based methods to generate realizable and optimized motion plans in complex environments. 
One recent example is H-MaP~\cite{cicek2024hmapiterativehybridsequential}, a hybrid sequential manipulation planner that decouples object trajectory planning from agent motion planning, leveraging informed contact sampling and waypoint generation to enhance robustness.
Another recent work addresses the challenges of sequential manipulation planning in the presence of obstacles and narrow passages by decomposing complex tasks into a sequence of easier pick-and-place subproblems~\cite{levit2024solvingsequentialmanipulationpuzzles}.
The approach integrates forward search with an optimization-based Task-and-Motion Planning (TAMP) solver~\cite{toussaint2015logic,toussaint2018differentiable}, allowing the agent to generate a feasible manipulation plan in spatially constrained environments more effectively.
Other TAMP-related hybrid approaches focus on modeling modes with grasps~\cite{Hauser2011, Schmitt2017, Kingston2020, Vega-Brown2020}. 
Although recent advancements in hybrid motion planning have improved performance in complex and constrained environments, generating robust and computationally efficient solutions remains challenging, particularly in identifying critical subgoals and key objects.

\subsection{Rearrangement Planning}
Rearrangement planning is a crucial aspect of sequential manipulation, particularly in environments where clutter obstructs the completion of a task.
~\cite{rearrangement_puzzle} introduces factored state spaces to model rearrangement puzzles, allowing for efficient problem decomposition where~\cite{zhang2024learn2decomposelearningproblemdecomposition} utilizes a learning-based approach to decompose long-horizon tasks into smaller, parallelized subproblems. 

The rearrangement planning has conceptual similarities with the Navigation Among Movable Objects (NAMO) and Manipulation Among Movable Objects (MAMO)~\cite{stilman2007manipulation}.
Recent advances have improved efficiency and adaptability through search-based planning~\cite{ren2024search}, affordance-driven strategies~\cite{wang2020affordance}, socially-aware object placement~\cite{renault2020modeling} and contact-aware motion planning via pushing and grasping~\cite{wang2025contact}.

Currently, few approaches integrate manipulation planning with obstacle relocation in constrained and complex environments.
Our work proposes a generalizable framework that can generate robust and computationally efficient manipulation plans in the presence of obstacles.

\section{Preliminaries}
In our proposed framework, we use a hybrid motion planning approach to solve maze-like tasks in cluttered environments based on tasks from \cite{levit2024solvingsequentialmanipulationpuzzles}.

\subsection{Optimization Based Motion Planning with KOMO}

The SeGMan algorithm employs k-order motion optimization KOMO \cite{14-toussaint-KOMO} for motion planning between subgoals generated by Bi-RRT \cite{Bi_RRT}.
\begin{equation}
\begin{aligned}
\label{eq:KOMO}
\min_{x_{0:T}} \quad & \sum_{t=0}^{T} f_t(x_{t-k:t})^T f_t(x_{t-k:t})\\
\text{s.t.} \quad & \forall t : g_t(x_{t-k:t}) \leq 0, \quad h_t(x_{t-k:t}) = 0,
\end{aligned}
\end{equation}
where $x_t \in \mathcal{X}$, and $\mathcal{X}$ is the configuration space for the scene, including agents with n-DoFs and $m$ movable objects. 
The set of $x_{t-k}$ represents consecutive k+1 states from $x_{t-k}$ to $x_{t}$. 
$f_t$,  $h_t$, and $g_t$ represent the cost function, equality, and inequality constraints, respectively, for timestep $t$. 

In KOMO, logical constraints are translated into geometric constraints that define the motion feasibility of the agent and objects.
In our work, we used logical constraints \textit{touch}, \textit{stable}, \textit{positionDiff}, and \textit{stableOn}, which correspond to the following geometric constraints in KOMO. 
The \textit{touch} constraint ensures that the Euclidean distance between the agent and the object's grasp point is zero, effectively modeling contact as an equality constraint. 
The \textit{stable} constraint ensures that an object remains static at specific timesteps by preventing changes in its position and orientation, which is also formulated as an equality constraint. 
The \textit{positionDiff} constraint maintains a bounded relative position between two objects or between the agent and an object, represented as an inequality constraint on the Euclidean distance between them. 
Finally, the \textit{stableOn} constraint ensures that an object remains stably placed on a surface by constraining its contact points and enforcing stability conditions, combining both equality and inequality constraints. 
These geometric formulations allow KOMO to optimize motion while satisfying stability, contact, and positioning requirements needed for executing not only a single but also a sequence of pick-and-place actions jointly, which is crucial for the problems we consider.

\subsection{Problem Formulation}
Our aim is to solve pick-and-place problems where the goal object, $o_{\text{goal}}$, should be placed in goal position, $g$, in a cluttered environment, often with narrow passages. 
We denote the initial configuration as $x_0$ and goal configuration as $x_{\text{goal}}$.
The problem involves a combination of trajectory optimization and sampling-based planning to ensure that the object can be reached and placed at the goal location while respecting physical constraints. 

\setlength{\belowcaptionskip}{-10pt}
\begin{figure}[t]
   \centering
   \resizebox{\columnwidth}{!}{\includegraphics{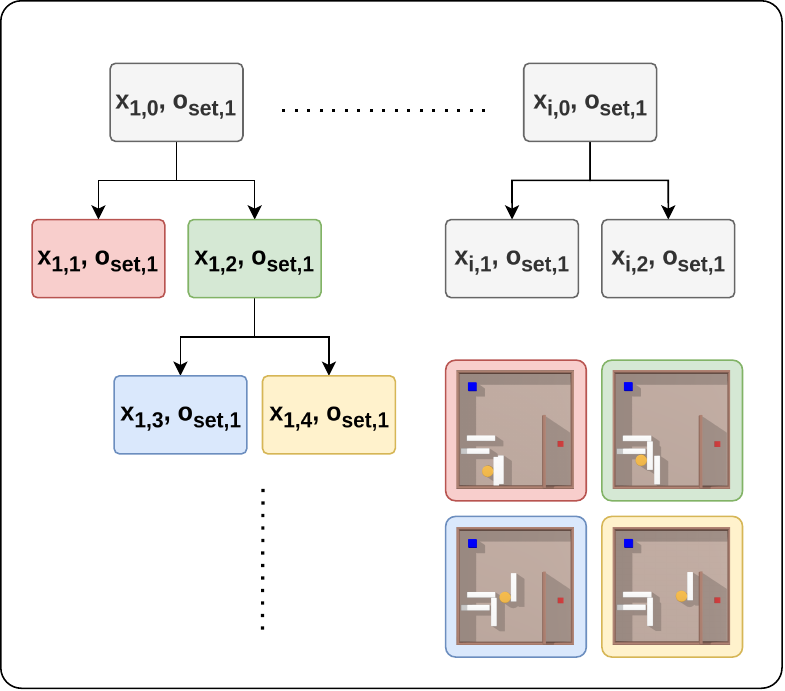}}
   \caption{The forest of $o_{set}$ trees.}
   \label{fig:forest}
 \end{figure}
 
Moreover, the environments might include movable objects that obstruct the path from the agent to the goal object $o_{\text{goal}}$ and from the goal object to the goal position, $g$. 
In this case, the agent needs to perform pick-and-place actions on these obstructing objects as well.
We denote movable objects (other than the goal object, $o_{\text{goal}}$) in the environment as $O=\{o_1,o_2,…,o_{m-1}\}$ for $m-1$ movable objects and the minimal subset containing the obstructing objects as $O_{\text{min}} \subseteq O$ such that removing all the elements of $O_{\text{min}}$ will result in a feasible plan $P$.
 
We divide the problem into several pick-and-place subproblems to reach the goal configuration. 
The final motion plan \( X \) is the concatenation of pick plans \( X^{o_p}_\text{pick}\left(t\right) \) and place plans \( X^{o_p}_\text{place}\left( t \right) \) for each subproblem \( p \). 
This can be mathematically represented as:
\[
X = \bigoplus_p \left( X^{o_p}_\text{pick}\left( t \right) \oplus X^{o_p}_\text{place}\left( t \right) \right)
\]
where:
\begin{itemize}
    \item \( X \) is the final motion plan consisting of a sequence of pick and place actions.
    \item \( X^{o_p}_\text{pick}\left( t \right) \) is the pick motion plan for object \( o_p \) as a function of timestep \( t \) for subproblem \( p \).
    \item \( X^{o_p}_\text{place}\left( t \right) \) is the place motion plan for object \( o_p \) as a function of timestep \( t \) for subproblem \( p \).
    \item \( \oplus \) denotes concatenation, meaning that the pick and the place plans are executed sequentially.
    \item \( \bigoplus_p \) represents the concatenation over all subproblems \( p \).
\end{itemize}

This problem can simply be stated as $X = X^{o_{\text{goal}} }_\text{pick}\left( t \right) \oplus X^{o_{\text{goal}} }_\text{place}\left( t \right)$ when there are no obstacles present in the environment. 
However, most of our tasks include obstacles where brute-force trial and error exponentially grows the problem of pick and place.
Therefore, we define the problem of obstacle subset, $O_{\text{set}}$, selection from the movable objects $O$ in the scene as follows.
We want to generate a $O_{\text{set}}$ such that $O_{\text{min}} \in O_{\text{set}}$, $O_{\text{set}} \subseteq \mathcal{P}(O)$ where $\mathcal{P}(O)$ is the power set of $O$, and $|O_{\text{set}}|$ is minimal.

{
\setlength{\belowcaptionskip}{-10pt}
\begin{figure*}[!th]
  \centering
  \resizebox{0.95\textwidth}{!}{\includegraphics{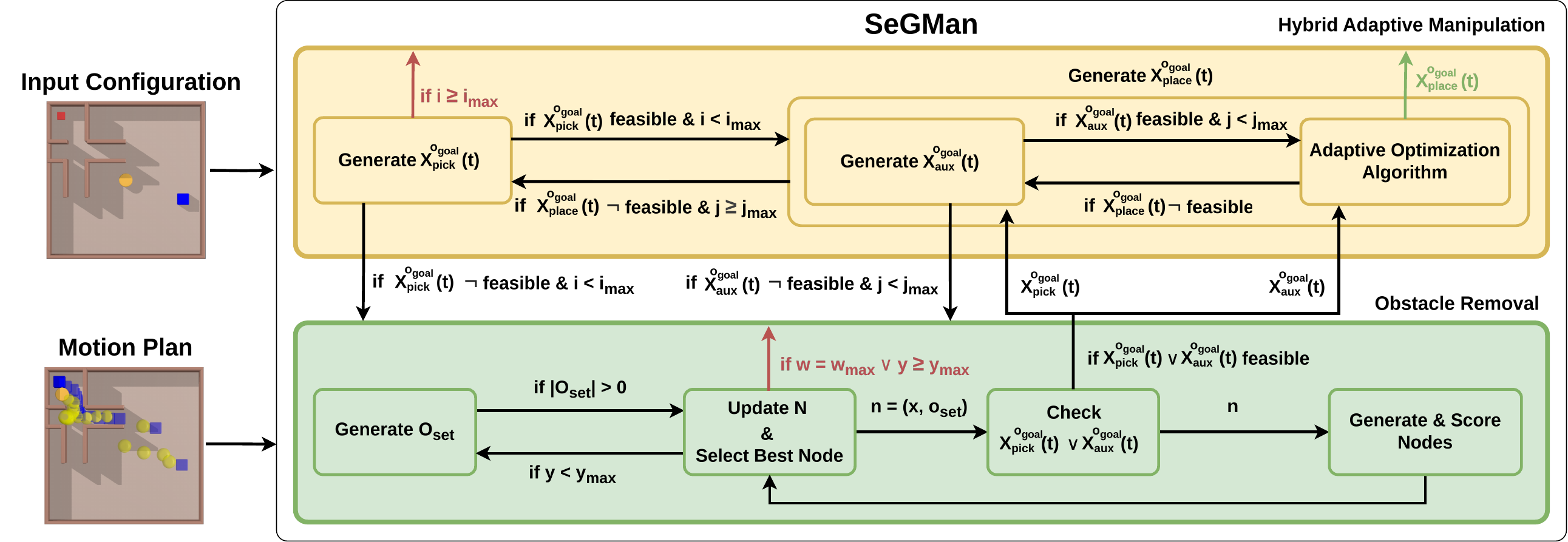}}
  \caption{System flowchart of SeGMan's two-phase architecture: (I) hybrid manipulation of goal object to goal position with adaptive subgoal selection, (II) forward search of determining obstacle objects and their removal. If no pick or place plan can be realized for the goal object, the obstacle removal phase clears the path. Once the paths are cleared, the framework continues to pick-place the goal object from where it left off.}
  \label{fig:framework}
\end{figure*}
}

\section{Sequential and Guided Manipulation Planner}
The framework given in Fig.~\ref{fig:framework} demonstrates the modules and how they interact with each other in the SeGMan algorithm. 
It consists of 2 main components: hybrid manipulation with adaptive subgoal selection and guided forward search.

\subsection{Hybrid Manipulation with Adaptive Subgoal Selection}
In this phase, a motion plan is generated for an object-goal tuple.
The motion plan $X$ consists of intermediate subgoals that are sequentially solved.
The object manipulation phase consists of pick-and-place sequences between the agent and $o_{goal}$ to generate an end-to-end motion plan. 

\subsubsection{Pick Plan}
First, a valid pick configuration, $x_{\text{pick}}^{o_{\text{goal}}} = X_{\text{pick}}^{o_{\text{goal}}}\left( T \right)$, where $T$ is the final timestep, is found using the non-linear constraint optimization explained in Eq. \ref{eq:KOMO}. 
Once a viable grasping configuration is found, a feasible trajectory $X_{\text{pick}}^{o_{\text{goal}}}\left( t \right)$ from $x_{0}$ to $x_{\text{pick}}^{{o_{\text{goal}}}}$ is generated with Bi-RRT.
If a feasible trajectory cannot be found from the initially proposed grasping configuration, new grasp configurations are tried until either a valid $X_{\text{pick}}^{o_{\text{goal}}}\left( t \right)$ is found or the trial limit is exceeded. 
If no feasible $X_{\text{pick}}^{o_{\text{goal}}}\left( t \right)$ is found, $x_{0}$ is checked for obstacles. 
If movable objects exist in $x_{0}$, their relocation is attempted using the methodology described in sec.~\ref{sec:obstacle removal}.

\subsubsection{Manipulation Plan}
Upon successful generation of  $X_{\text{pick}}^{o_{\text{goal}}}\left( t \right)$, a waypoint sequence from $o_{\text{goal}}$ to $g$ is generated.
Waypoint generation is performed in an auxiliary configuration $x_{\text{aux}, 0}^{o_{\text{goal}}}$ where the agent is removed and the $o_{\text{goal}}$ is treated as an agent. 
A trajectory $X_{\text{aux}}^{o_{\text{goal}}}\left( t \right)$ from $X_{\text{aux}}^{o_{\text{goal}}}(0)$ to $x_{\text{goal}}$ is computed using Bi-RRT.
If a feasible $X_{\text{aux}}^{o_{\text{goal}}}\left( t \right)$ cannot be found, the guided forward search explained in sec.~\ref{sec:obstacle removal} is utilized.

Once $X_{\text{aux}}^{o_{\text{goal}}}\left( t \right)$ is found, the intermediate configurations on the trajectory, ${x_{i}^{o_{\text{goal}}}}$, are used as a sequence of subgoals to generate $X_{\text{place}}^{o_{\text{goal}}}\left( t \right)$.
Rather than utilizing every ${x_{i}^{o_\text{goal}}}$ as a subgoal, the Adaptive Subgoal Selection Algorithm (Alg.~\ref{alg:adaptive}) iteratively selects closer subgoals within narrow passages, while maintaining efficiency in less constrained areas by selecting farther subgoals when feasible. 
If a feasible $X_{\text{place}}^{o_{\text{goal}}}\left( t \right)$ cannot be found within limited attempts, the trajectory is resampled with a finer resolution. 
If increasing the trajectory granularity still does not result in a feasible $X_{\text{place}}^{o_{\text{goal}}}\left( t \right)$, a different trajectory  $X_{\text{aux}}^{o_{\text{goal}}}\left( t \right)$ is generated.
This iterative process continues until either a $X_{\text{place}}^{o_{\text{goal}}}\left( t \right)$ is found or the iteration limit is reached, at which point the algorithm goes back to generating $X_{\text{pick}}^{o_{\text{goal}}}\left( t \right)$ where possible obstacles can be removed in the subsequent steps.
The planner stops when a final motion plan is found or if a maximum trial limit is reached.

\setlength{\belowcaptionskip}{-10pt}
\begin{algorithm}[!t]
\caption{Adaptive Subgoal Selection Algorithm}
\label{alg:adaptive}
\small
\DontPrintSemicolon
\SetAlgoNoLine
\SetKw{Break}{break}  
\KwIn{Goal object trajectory $X_{\text{aux}}^{o_{\text{goal}}}$, step incrementation threshold $\theta$}

\KwOut{Motion plan \(X_{\text{place}}^{o_{\text{goal}}}\)}
\textbf{Initialize:} \(step_\text{max} \gets |X_{\text{aux}}^{o_{\text{goal}}}| - 1\), \(p_i,step_i \gets step_\text{max}\), \(p_{prev} \gets 0\), \(\beta \gets 0\), \(X_{\text{place}}^{o_{\text{goal}}} \gets []\)\;
\While{\(o_{goal} \; \text{not reached} \; g\)}{
    \(x_{i}^{o_{\text{goal}}} \gets X_{\text{aux}}^{o_{\text{goal}}}[p_i]\)\;
    \(feasible \gets False\)\;

        \(X_{\text{place}}^{o_{\text{goal}}}, feasible \gets \textbf{KOMO}(x_{i}^{o_{\text{goal}}})\)\;
        \If{\(feasible\)}{
            \(\beta \gets \beta + 1\)\;
            \(p_{prev} \gets p_i\)\;
            \If{\(\beta \ge \theta\)}{
                 \(step_i \gets \min\{step_i \times 2,\, step_\text{max}\}\)\;
            } 
            \(p_i \gets \min\{p_i + step_i,\, step_\text{max}\}\)\;
        }
    
    \Else{
        \(\beta \gets 0\)\;
        \If{\(step_i = 1\)}{
            \Return{\(None\)}
        }
         \(step_i \gets \max\{1, step_i \ / \ 2 \}\)\;
         \(p_i \gets \min\{p_{prev} + step_i,\, step_\text{max}\}\)\;
    }
}
\Return{\(X_{\text{place}}^{o_{\text{goal}}}\)}
\end{algorithm}

\setlength{\belowcaptionskip}{-10pt}
\begin{figure}[t]
   \centering
   \resizebox{\columnwidth}{!}{\includegraphics{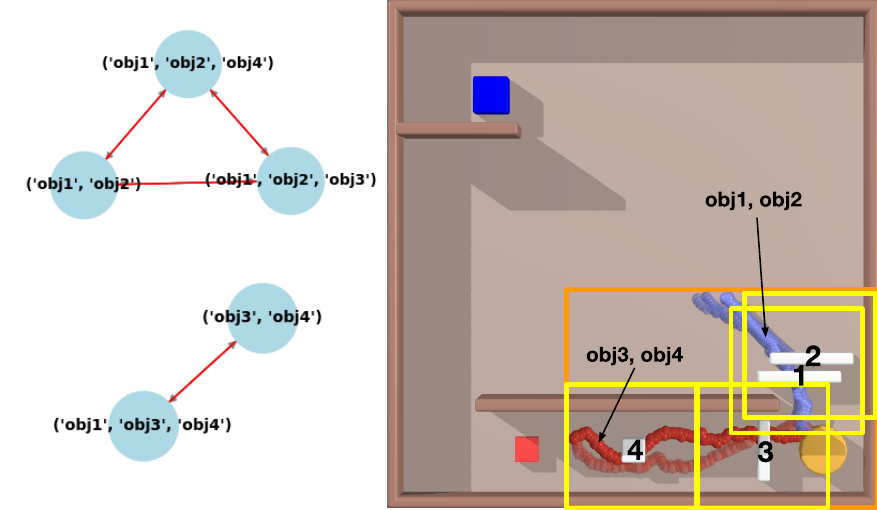}}
   \caption{Left: The clustered object sets (arrow direction is not important). Right: Two different clusters (red and blue) with each path are generated for $o_{set} \in C_i$. The more transparent the color of the path is, the less the $o_{set}$ is prioritized .}
   \label{fig:cluster}
 \end{figure}

\subsection{Obstacle Removal}
\label{sec:obstacle removal}
In this section, we first introduce our forward search strategy for obstacle removal. Then, we describe how the forward search is guided by the node selection policy and local occupancy grids (LOG). Finally, we explain how we create new nodes to expand the search.
\subsubsection{Forward Search}
If a clear path from the agent to the goal object, $o_{\text{goal}}$, or from the goal object to the goal position cannot be found, our framework performs a guided forward search over possible configuration-obstacle set tuples, $\left(x, o_{\text{set}}\right)$, which we call a \textit{node} $n \in N$.
Our method performs a search in a forest (see Fig.~\ref{fig:forest}), where each tree is associated with a different obstacle subset, $o_{\text{set}} \subseteq O$.
Nodes in a particular tree represent different configurations of the obstacles associated with this tree.
We generate $y$ number of candidate subsets using the subset selection policy (sec.~\ref{sec:object set generation}) and create a forest of $y$ number of trees.
We score each new node using a scoring policy and select the node with the highest score in the forest to expand (sec.~\ref{sec:node selection}).
When we are expanding a node, subgoal locations are generated for the tree's associated obstacle set (sec.~\ref{sec:node generation}), guiding the search through only a subset of objects.
After the existing $y$ trees reach a certain depth threshold, we add new trees for $y$ more subsets until all object sets are explored to balance the depth and breadth of the search.
The search is terminated when a node is found such that the configuration $x$ contains a feasible path or the maximum search limit is reached.

\setlength{\belowcaptionskip}{-10pt}
 \begin{figure}[t]
  \centering
  \resizebox{\columnwidth}{!}{\includegraphics{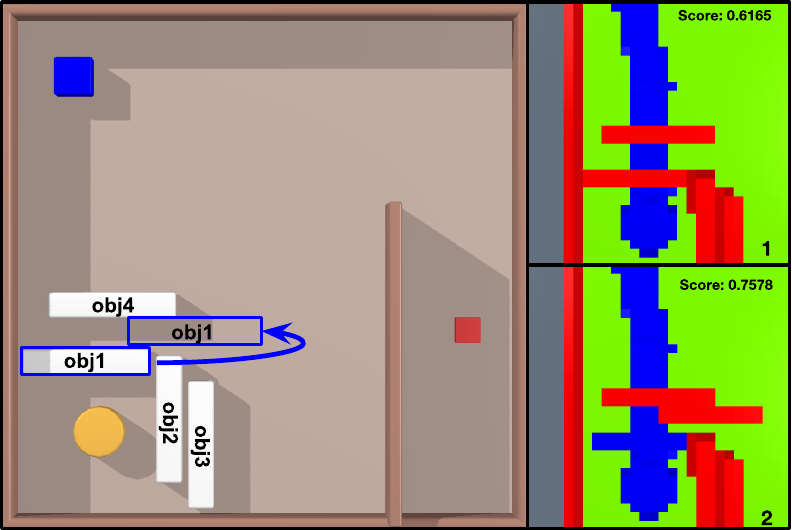}}
  \caption{Left: The Four Block scene, \textit{obj1} is relocated (blue arrow).  Right: Local occupancy grid (LOG) for \textit{obj4}. The colors red and green, respectively, represent objects and free spaces. The color blue represents $o_{set}$ trajectory, the initial positions of each object in \( o_{\text{set}} \), and the initial position of the agent. When \textit{obj1} moves, blue and green areas expand; thus, the configuration score increases.}
  \label{fig:ocugrid}
\end{figure}

{
\begin{figure*}[!th]
  \centering
  \resizebox{0.9\textwidth}{!}{\includegraphics{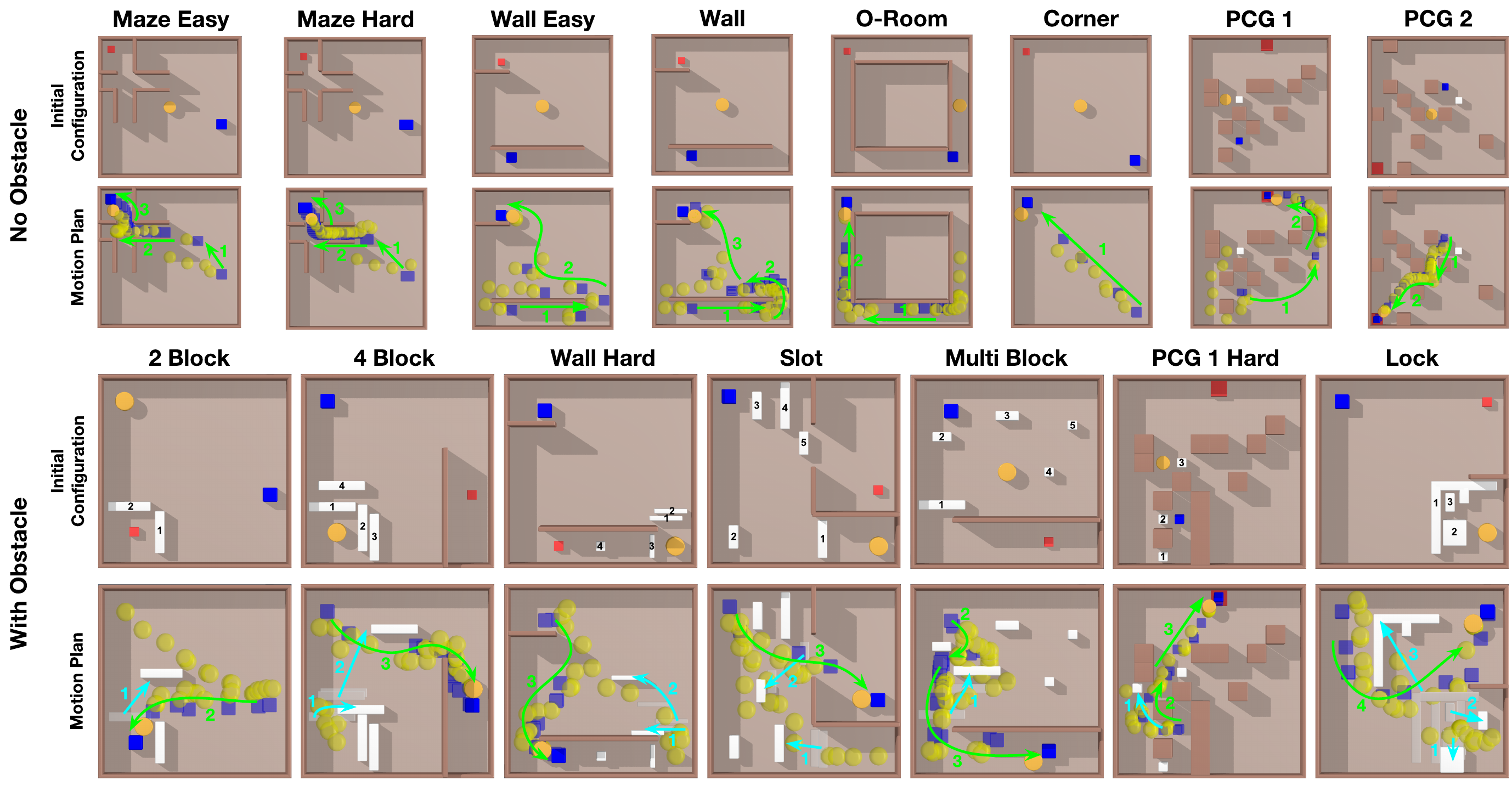}}
  \caption{Illustration of 15 tasks used to evaluate the proposed framework and baselines. The upper rows show the initial configuration of the scene, while the lower rows show the final motion plan trajectories. Green arrows indicate $o_{goal}$ movements, whereas turquoise arrows represent the relocation of objects $o \in O$.}
  \label{fig:scene}
\end{figure*}
}
\subsubsection{Object Set Generation}
\label{sec:object set generation}
In complex environments with numerous objects, it is important to guide the forward search toward a subset of objects rather than considering all movable objects. 
Preferably, object sets that include critical objects blocking the path should be selected.

For a subset $o_{\text{set}} \in \mathcal{P}(O)$ where $\mathcal{P}(O)$ is the power set of $O$, first a preliminary elimination is done by checking whether a clear path exists when the objects in $o_{\text{set}}$ are completely removed from the configuration.
If a solution does not exist, we eliminate this subset as a possible feasible subset.

After the preliminary elimination, similar sets are clustered based on the similarity of their generated paths, using Dynamic Time Warping (DTW) \cite{sakoe1978dynamic}. The sets containing redundant objects follow similar paths to the set with no redundant objects since redundant objects contribute little to path variation. However, they still might be clustered into different groups due to randomization in Bi-RRT.
To minimize variations introduced by Bi-RRT, we check the similarity of only a small path segment.
The path segment in the smallest region containing each $o \in O$ (Fig.~\ref{fig:cluster}) is used for similarity.
Finally, the clustered sets, $C=\{C_1,C_2,...,C_i\}$, are scored using \( S_\text{o} \) to prioritize sets that include fewer redundant objects.

Within the same cluster, the smallest subset contains the most critical objects, as other sets in the cluster are likely supersets that include redundant objects.
Thus, selecting only the smallest subset in each cluster might seem preferable to scoring all sets and prioritizing the smallest ones.  
However, in certain scenes (Lock problem in Fig.~\ref{fig:scene}), removing some redundant objects might be necessary to successfully relocate the objects in the smallest subset.
For this reason, we adopt a soft threshold with subset priority scoring rather than strictly selecting the smallest subset, as this approach enhances the generalizability of the proposed framework.

\subsubsection{Local Occupancy Grid (LOG)}
\label{sec:log}
LOG is proposed to guide the search toward generating configurations that are more likely to yield feasible \( X_{\text{pick}}^{o_{\text{goal}}}\left(t\right) \) or \( X_{\text{aux}}^{o_\text{goal}}\left(t\right) \). LOG is used for configuration scoring and subgoal generation.

For configuration scoring, LOG is generated for each \( o \in o_{\text{set}} \), localized around the initial position of \( o \).  
Fig.~\ref{fig:ocugrid} shows an example LOG, where red and green areas represent free space and objects, respectively.  
Blue areas correspond to the path taken by the agent if \( o \in o_{\text{set}} \) were removed, the initial position of each $o$, and the agent's initial position. 
The configuration score $S_{\text{x}}$ is obtained by calculating the green and blue areas where each color has a score per unit area defined with blue areas weighted twice as much.  
The $S_{\text{x}}$ prioritizes configurations that free the object's initial position without further obstructing the agent or its trajectory.  

The subgoals for node generation (sec.~\ref{sec:node generation}) are obtained from the LOG, which is localized around the object's current position in \( x_i \). The green areas in the LOG are proposed as candidate subgoal locations.

\subsubsection{Node Selection}
\label{sec:node selection}
We use the following node selection policy, which is:
\begin{equation}
\begin{aligned}
\arg \max_{n \in N} S(n) &= \alpha \cdot  \sqrt{\frac{1}{V_n}}  + \gamma^{V_n} \cdot {|C_i| \over \arg \max_{C_k \in C} |C_k|}\\
&\quad \cdot \left({1 \over |o_{\text{set}}|}  \cdot  S_{\text{r}}(n) \cdot  S_{\text{x}}(n) \right)^2 
\end{aligned}
\label{eq:Scoring Policy}
\end{equation}
where $o_\text{set} \in C_i$, $n=(x,o_{\text{set}})$ and $S(n)$ is the node score. Since our goal is to find the minimal feasible subset for object removal, we first expand the nodes that have smaller $o_{\text{set}}$ size, $|o_{\text{set}}|$. 
Therefore, the score is inversely proportional to the size of the subset.  
Moreover, the minimal subset is more likely to be in a larger cluster of subsets, $C_i$ since the supersets of the minimal subset will also yield feasible configurations; thus, such clusters receive higher scores.

$S_{\text{r}}$ is the reachability score.
For a node $n=(x,o_{\text{set}})$, $S_{\text{r}}(n) = r + 1$, where $r$ is the number of reachable objects $o \in o_{\text{set}}$ in configuration $x$. 
$S_{\text{x}}$ is the configuration score obtained using the local occupancy grid as explained in sec. \ref{sec:log}.
The remaining terms are used to ensure the exploration of nodes; $\alpha$ is the exploration factor, $V_n$ is the visit count of the current node $n$ and $\gamma$ is the discount factor.

\subsubsection{Node Generation}
\label{sec:node generation}
Node with the highest score, $n=(x, o_{\text{set}})$, is expanded by repositioning the reachable objects in $o_{\text{set}}$ to the proposed subgoal positions. 
The subgoals for each object are obtained from the LOG, and a randomly sampled subset of these subgoals is tested. 
The pick-and-place sequence is optimized using Eq.~\ref{eq:KOMO} for each subgoal. 

The \textit{positionDiff} logical constraint is given to the optimizer as a soft constraint in this phase, allowing it to position the object in a feasible location near the specified subgoal rather than at the exact position.
This increases the computational efficiency in cases where the exact subgoal location is not feasible.

If the motion plan is feasible, the reachability score \( S_{\text{r}} \) is calculated.
Finally, each new node is scored using \( S(n) \), sorted, and the best subset of generated nodes is added to the complete node set \( N \).

\setlength{\intextsep}{0pt}
\setlength{\tabcolsep}{2pt} 
\begin{table*}[h!]
\centering
\caption{The computation time, solution success rate, and solution quality (pick and place count) results for the SeGMan,~\cite{levit2024solvingsequentialmanipulationpuzzles} Random,~\cite{levit2024solvingsequentialmanipulationpuzzles} Bottleneck, and H-MaP~\cite{cicek2024hmapiterativehybridsequential} after running each task 10 times. ``-" indicates failure to solve the task after 1000 seconds. }
\scriptsize 
\resizebox{\textwidth}{!}{ 
\begin{tabularx}{\textwidth}{l *{16}{>{\centering\arraybackslash}X}}
\toprule
\multirow{2}{*}{Task Name} & \multicolumn{4}{c}{SeGMan} & \multicolumn{4}{c}{~\cite{levit2024solvingsequentialmanipulationpuzzles} Random} & \multicolumn{4}{c}{~\cite{levit2024solvingsequentialmanipulationpuzzles} Bottleneck} & \multicolumn{4}{c}{H-MaP~\cite{cicek2024hmapiterativehybridsequential}}  \\ 
\cmidrule(lr){2-5} \cmidrule(lr){6-9} \cmidrule(lr){10-13} \cmidrule(lr){14-17}
 & Mean (s) & Std Dev (s) & Solved (\%) & PnP Count & Mean (s) & Std Dev (s) & Solved (\%) & PnP Count & Mean (s) & Std Dev (s) & Solved (\%) & PnP Count & Mean (s) & Std Dev (s) & Solved (\%) & PnP Count \\ 
\midrule

Maze Easy & 2.66 & 0.41 & 100 & 15.9 & 18.59 & 24.07 & 100.0 & 3.2 & 190.47 & 276.03 & 100.0 & 3.8 & 0.30 & 0.03 & 100.0 & 53.0 \\

Maze Hard & 3.96 & 1.94 & 100.0 & 18.2 & -- & -- & -- & -- & -- & -- & -- & -- & 0.28 & 0.02 & 100.0 & 47.1 \\

Wall Easy & 1.63 & 0.74 & 100.0 & 9.3 & 21.18 & 7.41 & 100.0 & 3.0 & 71.44 & 10.61 & 100.0 & 3.0 & 0.39 & 0.04 & 30.0 & 75.33 \\

Wall & 4.00 & 3.53 & 100.0 & 11.6 & 395.57 & 194.30 & 100.0 & 5.3 & 360.78 & 218.87 & 90.0 & 5.56 & 0.51 & 0.43 & 90.0 & 58.78 \\

O-Room & 4.04 & 2.50 & 100.0 & 16.0 & 146.35 & 105.85 & 100.0 & 3.8 & 473.23 & 237.72 & 90.0 & 6.33 & 0.43 & 0.07 & 70.0 & 76.43 \\

Corner & 0.62 & 0.08 & 100.0 & 3.7 & 0.19 & 0.32 & 100.0 & 2.0 & 0.31 & 0.09 & 100.0 & 2.0 & 0.20 & 0.02 & 100.0 & 51.1 \\

PCG 1 & 5.49 & 4.95 & 100.0 & 11.9 & 57.39 & 36.22 & 100.0 & 3.1 & 125.05 & 33.69 & 90.0 & 3.67 & 0.45 & 0.07 & 100.0 & 63.6 \\

PCG 2 & 13.33 & 9.99 & 100.0 & 19.3 & 10.61 & 2.22 & 100.0 & 3.0 & 38.82 & 10.37 & 100.0 & 3.1 & 0.28 & 0.01 & 70.0 & 41.57 \\
\midrule
Two Blocks & 8.99 & 0.55 & 100.0 & 4.7 & 1.77 & 0.91 & 100.0 & 2.3 & 6.11 & 1.05 & 100.0 & 4.0 & -- & -- & -- & -- \\

Four Blocks & 46.73 & 22.23 & 100.0 & 10.1 & 877.18 & 0.0 & 10.0 & 5.0 & 684.81 & 0.0 & 10.0 & 8.0 & -- & -- & -- & -- \\

Wall Hard & 37.42 & 2.53 & 100.0 & 8.9 & -- & -- & -- & -- & -- & -- & -- & --  & -- & -- & -- & -- \\

Slot & 26.10 & 3.38 & 90.0 & 7.77 & 577.52 & 77.20 & 60.0 & 5.33 & 594.00 & 221.52 & 90.0 & 4.89 & -- & -- & -- & -- \\

Multi Block & 16.79 & 2.39 & 90.0 & 10.66 & 232.58 & 322.49 & 90.0 & 3.89 & 305.33 & 219.44 & 30.0 & 5.0 & -- & -- & -- & --  \\

PCG 1 Hard & 46.49 & 4.95 & 90.0 & 4.7 & 373.48 & 0.0 & 10.0 & 4.0 & 616.41 & 292.22 & 60.0 & 3.5 & -- & -- & -- & -- \\

Lock & 378.93 & 326.51 & 90.0 & 10.4 & 719.86 & 123.05 & 40.0 & 4.75 & 900.42 & 178.56.0 & 30.0 & 5.0 & -- & -- & -- & -- \\

\bottomrule
\end{tabularx}
}

\label{tab:runtime_comparison}
\end{table*}

\section{Experiments}
We evaluated SeGMan on 15 pick-and-place tasks (Fig.~\ref{fig:scene}), which are representative of complex and constrained environments.
The tasks are adapted and extended from~\cite{levit2024solvingsequentialmanipulationpuzzles}.
The agent's (yellow cylinder) task is to place the goal object (blue square) in the goal position (red square), interacting with movable objects (white squares) when required.
The tasks are divided into two categories: 
\begin{itemize}
    \item \textbf{No-obstacle tasks} The solution does not require interaction with movable objects.
    \item \textbf{With Obstacle tasks} The solution requires relocating key movable objects to find a feasible motion plan. 
\end{itemize}

The tasks require multiple pick-and-place sequences to manipulate objects through narrow passages while avoiding obstacles.
Choosing the right subgoals for the right objects is important as many objects in the tasks are either redundant or interdependent.
Thus, the tasks effectively represent the key challenges of sequential manipulation in constrained and complex environments.
We compare SeGMan with two versions of~\cite{levit2024solvingsequentialmanipulationpuzzles} and H-MaP~\cite{cicek2024hmapiterativehybridsequential}. 
\footnote{As the source code for~\cite{levit2024solvingsequentialmanipulationpuzzles} was not available, we re-implemented two variations of their algorithm. For~\cite{cicek2024hmapiterativehybridsequential}, we re-implemented the code in Python.} 

The primary evaluation metrics are solution success rate and computation time.
Additionally, the solution quality is assessed based on the number of pick-and-place sequences, where a higher count indicates lower quality.
Each experiment is repeated 10 times to ensure statistical significance. 
If the computation time exceeds 1000 seconds, the run is aborted and recorded as a failure.

\section{Results}
The success rate, computation time, and PnP (Pick-and-Place) count are given in Table~\ref{tab:runtime_comparison}. 
SeGMan successfully generated a feasible motion plan for all tasks in under 1000 seconds, while baseline methods struggled to solve some tasks within the given time limit.

\subsection{No Obstacle}
SeGMan successfully generated feasible motion plans for all ``No Obstacle" cases while demonstrating computational efficiency compared to the baselines.
Moreover, its sampling-based approach results in consistent subgoal generation across trials, as reflected in the low standard deviation in computation time. 

We observe that the PnP count is small for tasks with clear spaces, such as the Corner task, while the count increases as the task includes narrow passages, like the Maze task. This is due to the adaptive subgoal selection method; in narrow passages, subgoals are chosen in finer granularity to ease the manipulation. 

Both variations of~\cite{levit2024solvingsequentialmanipulationpuzzles} performed as expected, as most of these tasks were originally from their work.
They generated the highest quality solutions for all tasks within 1000 seconds, except for the Maze tasks.
Among all methods, H-MaP \cite{cicek2024hmapiterativehybridsequential} had the shortest computation time but struggled with robustness and solution quality.
While SeGMan and~\cite{cicek2024hmapiterativehybridsequential} share a similar sampling-based approach, SeGMan's adaptive subgoal selection method led to higher-quality solutions.
This is because~\cite{cicek2024hmapiterativehybridsequential} does not select a subset of the generated subgoals and utilizes all of them. 

\subsection{With Obstacle}
To solve the ``With Obstacle" tasks, SeGMan employs the guided forward search for obstacle relocation. 
Despite the increased complexity of the puzzles, SeGMan successfully found solutions in nearly all trials.
Compared to the baselines, the proposed framework generates significantly faster solutions. 
SeGMan consistently produced robust, computationally efficient solutions while maintaining relatively high solution quality.

Compared to ``No obstacle" tasks, the solutions of these tasks produce lower PnP counts. There are two reasons for this behavior. Firstly, the environments in this category include more movable objects in the configuration but have fewer narrow passages. Secondly, in the case of PCG 1 Hard, the solution quality is better than PCG 1, because Bi-RRT generates a path passing through clear spaces in PCG 1 Hard, whereas in PCG 1, the path goes through narrower spaces. 

The baseline results provide insight into the difficulty of generating a solution for these tasks and highlight the performance of SeGMan.
H-MaP's object relocation strategy could not find feasible subgoals for the obstacles; thus, no solution was found for the tasks. 

Similarly,~\cite{levit2024solvingsequentialmanipulationpuzzles} struggles to generate a solution to these tasks, failing to find solutions for 1 out of 7 cases. 
The primary reason \cite{levit2024solvingsequentialmanipulationpuzzles} struggles to generate solutions is that when there are many obstacles in the scene, subgoals are generated for every reachable obstacle, increasing the search space considerably while SeGMan selectively moves a subset of movable objects in the configuration.

The subgoal scoring heuristic in \cite{levit2024solvingsequentialmanipulationpuzzles} does not differentiate configurations with varying obstacle placements, rendering it less effective, especially when obstacles need to be relocated to critical positions, as seen in Wall Hard and Slot tasks.
While both SeGMan and~\cite{levit2024solvingsequentialmanipulationpuzzles} implement forward search, SeGMan's LOG and clustering-based heuristics effectively reduced the search space and prioritized subgoals that cleared obstacles, leading to shorter solution generation times.

\subsection{Limitations \& Discussion}
There are several limitations of our work. Firstly, the proposed heuristics guide the forward search toward feasible configurations; however, there is a computational overhead in simpler cases.
In scenarios like Corner and Two Blocks,~\cite{levit2024solvingsequentialmanipulationpuzzles} outperforms SeGMan in most metrics, benefiting from its underlying TAMP framework.
Moreover, while the sampling-based subgoal generation increases robustness, the solution quality decreases due to the size of the subgoal set.
Although the subgoals are selected adaptively, the generated end-to-end motion plan is not necessarily optimal in solution quality.

The results reveal a trade-off between computation time and solution quality, with~\cite{levit2024solvingsequentialmanipulationpuzzles} performing better in solution quality, while H-MaP is computationally faster.
SeGMan offers a balanced approach, producing robust motion plans while offering efficient computation performance and relatively high solution quality.

\section{Conclusion}
In this work, we introduced SeGMan, a hybrid motion planner that enhances sequential manipulation in constrained, cluttered environments. By combining a hybrid motion planning method and heuristic-guided forward search, SeGMan efficiently identifies critical subgoals and generates robust motion plans.

Our approach addresses key challenges such as obstacle selection, subgoal feasibility, and fine object manipulation, where existing methods often struggle. Adaptive subgoal selection and iterative obstacle relocation with guided forward search enable generalization across environments. Empirical results show that SeGMan outperforms baselines, achieving a strong balance between computation time and solution quality.

SeGMan provides a framework that can be implemented in other domains, such as mobile robots and prehensile manipulation. Future work will focus on extending SeGMan to higher-dimensional domains and improving adaptability to real-world uncertainties, contributing to more intelligent and autonomous robotic planning.

\bibliographystyle{ieeetr}
\bibliography{main}
\vspace{12pt}

\end{document}